# Deep Multi-Scale Resemblance Network for the Sub-class Differentiation of Adrenal Masses on Computed Tomography Images

Lei Bi[a,*], Jinman Kim [a,*], Tingwei Su[b], Michael Fulham[a,c,d], David Dagan Feng[a,e], and, Guang Ning[b]

[a] School of Computer Science, University of Sydney, NSW, Australia

[b] Department of Endocrine and Metabolic Diseases, Ruijin Hospital, Shanghai Jiao Tong University School of Medicine, Shanghai, China

[c] Department of Molecular Imaging, Royal Prince Alfred Hospital, NSW, Australia

[d] Sydney Medical School, University of Sydney, NSW, Australia

[e] Med-X Research Institute, Shanghai Jiao Tong University, Shanghai, China

[*] Corresponding author: lei.bi@sydney.edu.au; jinman.kim@sydney.edu.au

*Abstract* — Objective: The accurate classification of mass lesions in the adrenal glands ('adrenal masses'), detected with computed tomography (CT), is important for diagnosis and patient management. Adrenal masses can be benign or malignant and benign masses have varying prevalence. Classification methods based on convolutional neural networks (CNNs) are the state-of-the-art in maximizing inter-class differences in large medical imaging training datasets. The application of CNNs, to adrenal masses is challenging due to large intra-class variations, large inter-class similarities and imbalanced training data due to the size of the mass lesions. Methods: We developed a deep multi-scale resemblance network (DMRN) to overcome these limitations and leveraged paired CNNs to evaluate the intra-class similarities. We used multi-scale feature embedding to improve the inter-class separability by iteratively combining complementary information produced at different scales of the input to create structured feature descriptors. We augmented the training data with randomly sampled paired adrenal masses to reduce the influence of imbalanced training data. Results: We used 229 CT scans of patients with adrenal masses for evaluation. In a five-fold cross-validation, our method had the best results (89.52% in accuracy) when compared to the state-of-the-art methods ($p < 0.05$). We conducted a generalizability analysis of our method on the ImageCLEF 2016 competition dataset for medical subfigure classification, which consists of a training set of 6,776 images and a test set of 4,166 images across 30 classes. Our method achieved better classification performance (85.90% in accuracy) when compared to the existing methods and was competitive when compared with methods that require additional training data (1.47% lower in accuracy). Conclusion: Our DMRN sub-classified adrenal masses on CT and was superior to state-of-the-art approaches.

## 1. Introduction

A large variety of abnormalities can be detected in the adrenal glands on abdominal / lower thoracic computed tomography (CT) scans. These abnormalities range from benign cystic changes / calcification to high grade malignant tumors that may arise in the gland itself (primary tumors) or reflect metastatic disease from another site e.g. a primary lung, bowel or skin cancer. The prevalence of mass lesions in the adrenal glands ('adrenal masses') is unclear but it has been suggested that the prevalence of adrenal adenomas, a sub-class of benign primary tumors, is 7% in subjects over 70 years of age [1]. Adrenal masses can be asymptomatic and there are a number of well-described clinical syndromes that are associated with tumors that secrete excess amounts of adrenal hormones such as cortisol, aldosterone, norepinephrine etc. [2-4]. An imaging specialist uses characteristics such as size, shape, homogeneity, morphology, density, presence of fat / calcification and characteristics of contrast enhancement on CT to sub-classify adrenal masses. There are large intra-class variations and large inter-class similarities between the various adrenal masses. The visual distinction between different classes can be subtle and texture variations within the same sub-classes can be marked. Thus assessment requires an experienced reader and poses challenges for less experienced imaging specialists and clinicians. Prior studies suggest that an automated computer aided diagnosis system (CAD) could improve accuracy and reduce image reading time [5, 6]. Hence our aim was to explore the possibility of developing an automated process to sub-classify adrenal masses that are assessed by a large endocrine service.

The main sub-classes of adrenal masses are: (i) adrenocortical carcinomas (ACAs); (ii) non-functional adrenal adenomas (NAAs); (iii) ganglioneuromas (GAs); (iv) adrenal myelolipomas (AMs); and (v) pheochromocytomas (PCCs). These adrenal masses vary in prevalence. Adenomas are very common as outlined above and carcinomas are rare. The normal adrenal glands are usually located in the posterior upper abdomen, antero-superior to the upper pole of each kidney. In Fig. 1 we show paired transaxial image CT slices of typical examples to illustrate the heterogeneity across and, within, these adrenal masses. There are differences in shape, density and patterns of contrast enhancement. AMs have a mixed soft tissue and fat density, vary in size and do not enhance. The varying size of different adrenal masses means that any training characteristics will be imbalanced. Existing classification methods tend to overfit to sub-classes with a large number of training characteristics and perform poorly on other sub-classes.

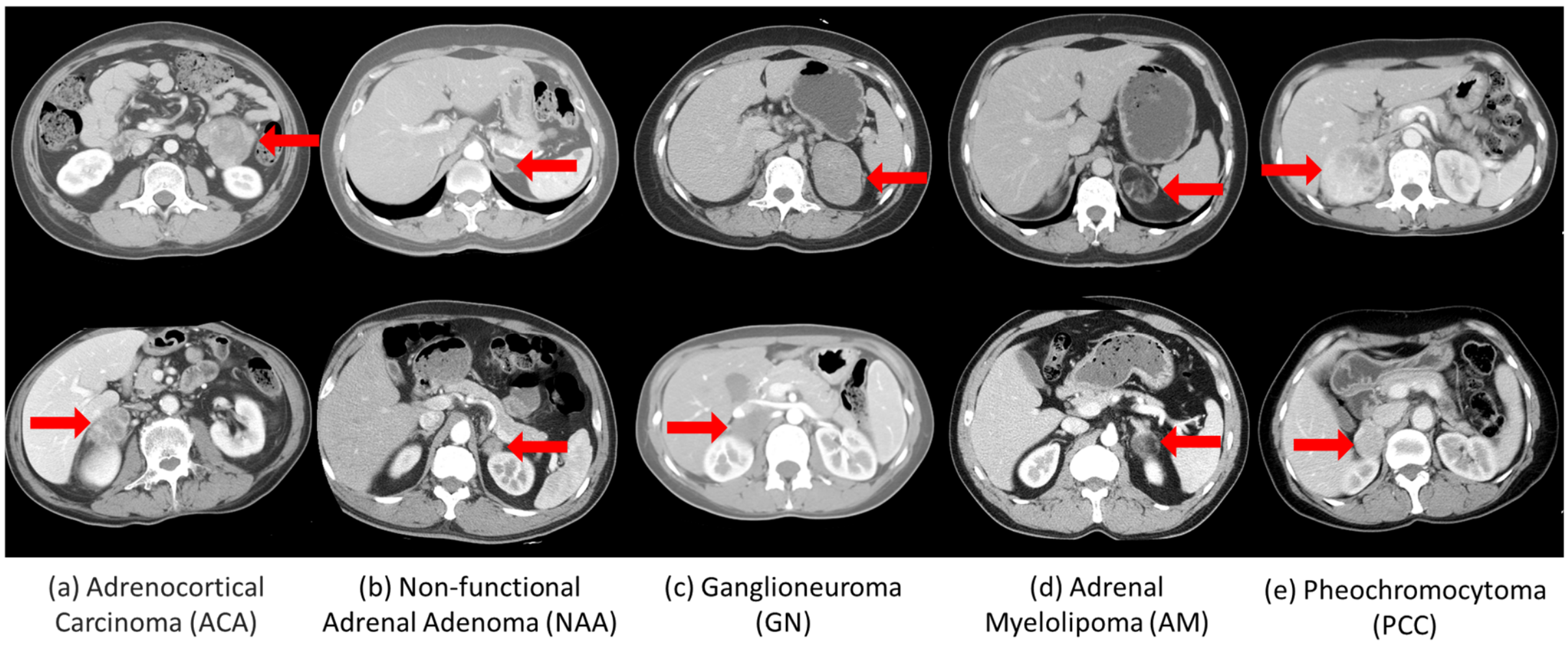

Figure 1. Paired transaxial contrast-enhanced CT images of different adrenal masses (red arrows).

### *1.1. Related Work*

Current automated medical imaging classification methods are: (i) traditional, using handcrafted features, with conventional classifiers; and (ii) deep learning using deep convolutional neural networks (CNNs). In traditional methods handcrafted techniques encode image characteristics as the feature descriptors and label image categories with supervised approaches. The most commonly used features include the local binary pattern (LBP) [6, 7], histogram of oriented gradients (HOG) [6, 8], gray level co-occurrence matrix (GLCM) and run length-based (RLE) texture [9-13] and wavelets [14, 15]. The extracted visual features train classifiers such as the support vector machine (SVM) [6, 8, 16, 17], K-nearest neighbor (KNN) [7, 18, 19] and random forest (RF) [14, 20-22]. Their performance, however, depends on effective pre-processing to reduce noise, tune a large number of parameters and manipulate the hand-crafted features and this limits their generalizability.

Deep learning uses CNNs to leverage large datasets to learn the features that best correspond to the appearance / semantics of the images [23]. CNNs can be trained in an end-to-end manner for efficient inference, i.e., images are taken as inputs and classification results are directly outputted. Anthimopoulos et al used a CNN with 5 convolutional layers to classify interstitial lung disease on chest CT [24]. Dou et al reported a cascaded 3D CNN to detect microbleeds on Magnetic Resonance (MR) scans [25]. Standard CNNs have difficulties when there are large intra-class variations, large inter-class similarities and imbalanced data [26, 27].

Many researchers have attempted to solve the problem of imbalanced data through augmenting the training dataset with additional image features or CNNs. Zhen et al proposed augmentation with Fisher features for melanoma detection on dermoscopy images [28]. Ahn et al leveraged pre-trained Visual Geometry Group (VGG) network features (trained on natural images e.g., ImageNet) with scale-invariant feature transforms [29]. Wang et al reported a multiscale rotation-invariant CNN with Gabor features for lung texture classification on chest CT [26]. Kumar et al used a combination of features from multiple CNNs [30]. Frid-Adar et al leveraged generative adversarial networks (GANs) to synthesize CT images for liver lesion classification [31]. Augmentation methods balance the data distribution by leveraging additional image features derived from using additional image datasets, image synthesis and feature hand-crafting. However, additional image datasets from the same domain may not be available and image datasets from a different domain e.g., natural images, have limited relevance to the target problem. Image synthesis may help to learn new features, however, the synthesized images can contain artifacts and noise, and these will have negative impact for augmentation. Feature hand-crafting methods usually require priori knowledge, which may limit generalizability to different datasets.

A number of researchers have modeled the inter-class differences to manage large intra-class variations and inter-class similarities in image datasets. Zhang et al trained multiple CNNs in a competitive manner to learn differences in the input data from different classes [27]. Ahn et al employed a convolutional sparse kernel network to learn class-specific image features for better discrimination [32]. Zhang et al proposed an attention residual learning network to learn subtle inter-class image features [33]. All these methods were designed to learn inter-class differences through maximizing inter-class distances, without minimizing large intra-class variations. Hence, they are less effective for images with large intra-class variations.

### 1.2. Our Contribution

We suggest a different approach to differentiate adrenal masses on CT and our contribution is as follows:

(1) We propose a similarity loss to derive features that is able to tolerate large intra-class variations. We leverage paired CNNs to evaluate the intra-class feature differences and then use the similarity loss to gradually learn the intra-class training samples with large feature distances.

(2) We propose a multi-scale feature embedding to refine feature learning. We progressively integrate the complementary feature representations produced at different scales (levels) of the input adrenal masses to create structured feature descriptors to then improve the discriminating power for sub-class classification.

(3) We leverage paired inputs to ameliorate imbalanced training data by augmenting the training data with randomly sampled data from different sub-classes to create the paired data. Thus all sub-classes have equal contributions to the final model and the risk of overfitting is minimized.

In the rest of the paper, we have Methods and Evaluation in Section 2, Results in Section 3, Discussion in Section 4 and Conclusions in Section 5.

## 2. METHODS

### 2.1. Materials

Our dataset comprised 229 contrast-enhanced abdominal CT scans of patients from the Department of Endocrine and Metabolic Diseases, Ruijin Hospital, Shanghai, China which is a major referral hospital for Endocrine disease in China. The studies were acquired on various scanners (GE LightSpeed and Discovery, Philips iCT, and Siemens SOMATOM) with CT resolution varying between 0.5625×0.5625 $mm^2$ to 0.9766×0.9766 $mm^2$ and transaxial slice thickness from 1.25mm to 5mm. The adrenal masses were separated into: (1) ACAs (n=54); (2) NAAs (n=35); (3) GNs (n=58); (4) AMs (n=33); and (5) PCCs (n=49). This radiologic assessment was done by an experienced imaging specialist. The assessment was not made on the basis of pathology but rather the imaging features on CT and blood tests. This is because biopsies of the adrenal glands are not usually carried out due to the location of the adrenal glands and their proximity to adjacent structures. PCCs are readily identified through the combination of serum hormone levels, clinical findings and imaging; the other sub-classes have characteristic appearances on CT. A senior clinician manually annotated the adrenal masses and a 3D bounding box was placed over each mass. The 3D bounding box was then separated into 2D bounding boxes based on transaxial slices and the 2D bounding box (image patch) was resized into 224 × 224. We followed the CT soft-tissue window and used Hounsfield units (HU) range of [-160, 240] to clip the CT images, which was then used as the input for training and testing. Table 1 summarizes the dataset; the imaging dataset was imbalanced with fewer NAAs, AMs and PCCs and fewer image slices in these sub-classes.

Table 1. A summary of the imaging dataset.

| # | ACA | NAA | GN | AM | PCC | Total |
|---|---|---|---|---|---|---|
| **Studies** | 54 | 35 | 58 | 33 | 49 | 229 |
| **Transaxial Slices** | 1707 | 245 | 1047 | 350 | 716 | 4065 |

## *2.2. Deep Multi-Scale Resemblance Network (DMRN)*

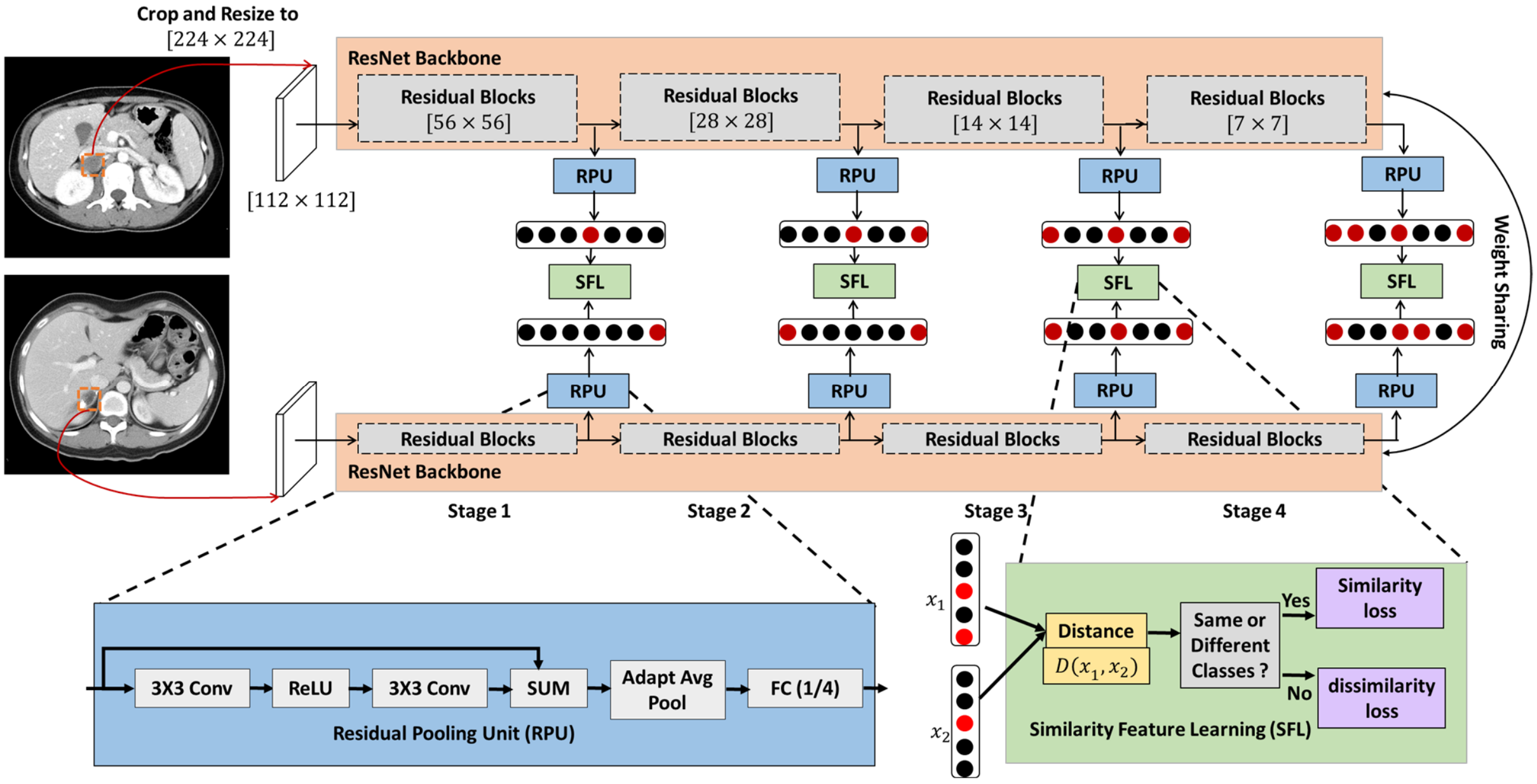

Figure 2. Flow diagram of our deep multi-scale resemblance network (DMRN). [·] indicates the output size.

A flow diagram of our method is shown in Fig. 2. In our design, we used the widely used ResNet (deep residual network) [34, 35] as the backbone to our DMRN method (more details of ResNet are provided in Section 2.2.1). The ResNet was applied to the paired training images (randomly produced) to generate feature maps at various scales. The feature maps from each scale were then embedded into paired feature vectors $x_1$and $x_2$ via the residual pooling unit (RPU). Then the similarity feature learning module measured the similarities between the two feature vectors at different scales and was trained to determine if the two feature vectors were from the same class ($Y = 0$ if $x_1$ and $x_2$ were from same class, and $Y = 1$ if they were from different classes).

### *2.2.1. Multi-Scale Feature Embedding*

A ResNet was used as the backbone for the initial design for its wide applications and scalability [34, 35]. The ResNet architecture has a number of residual blocks and a residual block uses a skip connection to bypass a few convolutional, batch normalization and rectified linear unit (ReLUs) layers at a time. The use of skip connections enables to reuse of activations from a previous layer (previous residual block) which thereby minimizes the problem of vanishing gradients.

Our ResNet has 4 stages representing 4 different feature maps – 56×56, 28×28, 14×14, and 7×7. For the output feature maps, we used an RPU to pool 2D feature maps into a single 1D feature vector and to pool spatial and semantic features at that stage. The RPUs start with two convolutional layers - a ReLU and summation layers with a kernel size of 3×3, which is a simplified version of the residual block in the original ResNet, where the batch normalization layers were removed. We applied an adaptive average pooling operation to aggregate the spatial information within the feature maps, where we used adaptive average pooling to the summed feature maps to down-sample the feature maps into 1×1. At the final stage of the RPU, we used a fully connected layer (FC) to use the inter-channel relationship and to reduce parameter overhead, where the output of the FC was set to 25% of the input in size.

### *2.2.2. Similarity Feature Learning*

Our similarity feature learning module learns subtle feature differences of the paired inputs at individual stages. Let $x_1^t$, $x_2^t$ be a pair of feature vectors derived from the fully connected layer of the two RPUs at stage $t$. Let $Y$ be a binary label assigned to this pair, where $Y = 0$ if $x_1^t$ and $x_2^t$ are from same adrenal mass sub-class, and $Y = 1$ if they are from different adrenal mass sub-classes. $D^t$ is the parameterized distance function of $x_1^t$ and $x_2^t$, which is defined as:

$$D^t(x_1^t, x_2^t, \boldsymbol{\theta}) = \|(x_1^t) - (x_2^t)\|_2 \tag{1}$$

Where $\boldsymbol{\theta}$ represents the network parameters. Then the overall loss function across all the stages is to find an optimal $\boldsymbol{\theta}$ can satisfy the following:

$$\mathcal{L} = arg\min_{\boldsymbol{\theta}} \sum_{t=1}^{T} \sum_{i=1}^{P} L^t(\boldsymbol{\theta}, (Y, x_1^t, x_2^t)^i) \tag{2}$$

Where $i$ represents the $i$-th training pair out of $P$ training sample pairs and $T$ represents different stages.

$L^t(\boldsymbol{\theta}, (Y, x_1^t, x_2^t))$ can be defined as:

$$L^t(\boldsymbol{\theta}, (Y, x_1^t, x_2^t)) = (1 - Y)L_S^t(D^t) + YL_D^t(D^t) \tag{3}$$

$L_S^t$ represents the similarity loss and $L_D^t$ is the dissimilarity loss, which are defined as:

$$L_S^t = \frac{1}{2}(D^t(x_1^t, x_2^t, \boldsymbol{\theta}))^2 \tag{4}$$

$$L_D^t = \frac{1}{2}\{\max(0, m - D^t(x_1^t, x_2^t, \boldsymbol{\theta})\}^2 \tag{5}$$

where $m$ is a margin threshold and defines when the $x_1^t$ and $x_2^t$ from different sub-classes will contribute to the loss. Based on the work proposed by Hadsell et al [36], we set $m = 1$.

*2.2.3. Training and Inference*

We trained the DMRN in an end-to-end manner by minimizing the overall loss between the predicted results $\boldsymbol{X}$ (where $x_1, x_2 \in \boldsymbol{X}$) and the binary annotation $\boldsymbol{Y}$ ($Y \in \boldsymbol{Y}$) of the paired data. The network parameters were then iteratively updated using the stochastic gradient descent (SGD) algorithm [37]. We used a weight sharing strategy to ensure that both branches were updated simultaneously. Exhaustive pairing results in $P = \frac{1}{2} \times f \times (f - 1)$ number of pairs, where $f$ is the number of annotated training data. We balanced the computation time and feature learning outcomes by using a random sampling strategy. For each annotated training image, we randomly selected another annotated image to create a pair and results in $P = f$ number of pairs. At the inference stage, we applied one branch of the DMRN to the input image for feature embedding, where we used the fully connected layer of the last stage of the DMRN to extract features. The extracted features were then used to estimate a probability score corresponding to the input image depicting one of the adrenal mass sub-classes. We used a support vector machine (SVM) [38] with a linear kernel as the classifier, trained with the same feature embedding process. A linear kernel took less than 20 seconds to train the SVM.

*2.3. Implementation Details*

We used a 2D based CNN for its GPU memory efficiency. We averaged the output probabilities for all transaxial slices for each study as the final output for each scan. Off-the-shelf PyTorch version of the 101 layer ResNet trained on the ImageNet dataset was used as a generic pre-trained model to guide the training initialization of our data [39].

It took about 10 hours to fine-tune over 200 epochs with a batch size of 10 on a 11 GB Nvidia 2080 Ti GPU.

### *2.4. Experimental Setup*

We used a five-fold cross-validation to evaluate our method. Specifically, we randomly divided the 229 studies into 5 distinct training and test sets for the 5-fold cross-validation evaluation protocol. We ensured that images from each study could only belong to the training set or the test set. We compared the performance of the DMRN to methods that are regarded as 'state-of-the-art' for medical image classification. These 'state-of-the-art' methods included: (1) Synergy based deep learning model (SDL) [27] – multiple CNNs were used to learn the differences of the input data of different classes in a competitive manner. (2) Attention residual learning network (ARL) [33] – ARL uses attention module to focus on learning subtle inter-class image features. (3) VGG [40] – A 19 layer VGG network. (4) Traditional handcrafted features (HC) with SVMs for classification and we followed existing methods [6] to use LBP and HOG features as the HCs. We used a patch size of 64 for LBP feature and a patch size of 32 for HOG feature. This resulted in a 531-d LBP feature and a 1296-d HOG feature. The extracted LBP and HOGs were concatenated into a single feature vector and were then used for classification. (5) A 3 dimensional convolutional neural network (3D-CNN). A similar approach was proposed by Dou et al [25] for microbleed detection on MR images. We used 6×3D convolutional layers followed by 3 fully connected layers with cross-entropy loss for classification. We also added 3D batch normalization layers and ReLU layers after each 3D convolutional layers. (6) ResNet-50, ResNet-101, and ResNet-152 – ResNet with 50, 101 and 152 layers [34].

We also conducted ablation experiments to evaluate the contributions of individual components of our approach and this included: (1) DMRN (w.o. MS) – our method without using multi-scale feature embedding (MS); (2) DMRN (w.o. MS w.o. SFL) – our method without using MS and similarity feature learning (SFL). We replaced the similarity loss with a binary cross entropy loss function. Specifically, we concatenated the output feature vectors $x_1$and $x_2$. We then passed the concatenated feature vector to a fully connected layer. The output of the fully connected layer was applied with a Sigmoid activation function, which was then compared to $Y$ ($Y = 0$ if $x_1$and $x_2$ were from same class, and $Y = 1$ if they were from different classes). and (3) DMRN (w.o. RS w.o. MS w.o. SFL) – our method without using MS, SFL and random sampling (RS).

### 2.5. Evaluation Metrics

We used the evaluation metrics: accuracy (Acc.), sensitivity (Sen.), specificity (Spe.), precision (Pre.) and F1 score (F1).

$$Acc. = \frac{|TP|+|TN|}{|TP|+|TN|+|FP|+|FN|} \tag{6}$$

$$Sen. = \frac{|TP|}{|TP|+|FN|} \tag{7}$$

$$Spec. = \frac{|TN|}{|TN|+|FP|} \tag{8}$$

$$Pre. = \frac{|TP|}{|TP|+|FP|} \tag{9}$$

$$F1 = \frac{2\cdot|TP|}{2\cdot|TP|+|FP|+|FN|} \tag{10}$$

where *TP* is the true positive, *TN* is the true negative, *FN* is the false negative and *FP* is the false positive. All the evaluation metrics were calculated for individual classes as a one-versus-all approach except for accuracy.

## 3. Results

### 3.1. Classification

Table 2 shows the overall results and Table 3 details the results on individual sub-classes. Both Tables show that the DMRN had the best overall performance across all the measurements and outperformed the second-best method with >4% in accuracy. We have also conducted paired-sample t-test of our results to the top performing comparison methods; with a $p < 0.05$ implying that the improvements are significant.

Fig. 3 shows classification results of randomly selected challenging studies with classification probabilities (confidence). It shows that the DMRN was the only method that produced the correct classification with the highest confidence.

Table 4 and Table 5 show the multi-class classification results on the transaxial slices. The lower classification results in Table 4 and Table 5 compared to Table 2 and Table 3 are expected. This is attributed to the fact that some transaxial slices e.g., the first and last slices of the adrenal mass may not have sufficient lesion information for characterization. Nonetheless, our method had large improvement to the existing methods, with consistent improvement in classification results across all the classes. These findings are consistent with the findings found in the lesion level classification results.

Table 2. Comparison of multi-class classification results (lesion level), where **red** represents the best, and **blue** represents the second-best results.

| | Acc. | Sen. | Spe. | Pre. | F1 | $p$-Value |
|---|---|---|---|---|---|---|
| **HC** | 57.64 ± 8.36 | 61.05 ± 10.01 | 89.22 ± 1.33 | 57.64 ± 7.49 | 59.00 ± 9.70 | $p < 0.01$ |
| **3D-CNN** | 63.32 ± 7.11 | 62.64 ± 5.67 | 90.61 ± 1.57 | 65.89 ± 6.46 | 63.52 ± 6.73 | $p < 0.01$ |
| **VGG** | 79.91 ± 7.36 | 81.47 ± 7.30 | 94.95 ± 1.84 | 81.02 ± 7.31 | 80.78 ± 7.43 | $p < 0.01$ |
| **ResNet-50** | 81.22 ± 9.17 | 82.74 ± 8.34 | 95.17 ± 2.42 | 83.96 ± 5.60 | 82.39 ± 8.55 | $p < 0.01$ |
| **ResNet-101** | 82.10 ± 11.60 | 82.48 ± 12.62 | 95.37 ± 2.92 | 84.39 ± 7.17 | 83.02 ± 13.34 | $p < 0.01$ |
| **ResNet-152** | 83.41 ± 6.33 | 83.60 ± 5.99 | 95.73 ± 1.57 | 84.93 ± 5.82 | 84.06 ± 5.17 | $p < 0.01$ |
| **ARL** | 82.53 ± 6.89 | 83.59 ± 5.70 | 95.54 ± 1.77 | 83.15 ± 6.57 | 83.27 ± 6.64 | $p < 0.01$ |
| **SDL** | 84.72 ± 5.08 | 86.03 ± 4.62 | 96.15 ± 1.27 | 85.18 ± 5.07 | 85.38 ± 4.87 | $p < 0.05$ |
| **DMRN** | 89.52 ± 6.26 | 90.73 ± 5.58 | 97.34 ± 1.53 | 89.91 ± 6.36 | 89.96 ± 6.27 | - |

Table 3. Comparison of multi-class classification results (lesion level) for individual sub-classes.

| | | VGG | ResNet-50 | ResNet-101 | ResNet-152 | ARL | SDL | DMRN (Ours) |
|---|---|---|---|---|---|---|---|---|
| **Sen.** | **ACA** | 79.63 | 90.74 | 90.74 | 90.74 | 83.33 | 83.33 | 92.59 |
| | **NAA** | 88.57 | 88.57 | 80.00 | 77.14 | 77.14 | 88.57 | 94.29 |
| | **GN** | 70.69 | 74.14 | 79.31 | 77.59 | 70.69 | 77.59 | 81.03 |
| | **AM** | 90.91 | 96.97 | 90.91 | 90.91 | 96.97 | 96.97 | 100.00 |
| | **PCC** | 77.55 | 63.27 | 71.43 | 81.63 | 89.80 | 83.67 | 85.71 |
| **Spe.** | **ACA** | 94.29 | 87.43 | 92.57 | 92.57 | 93.71 | 95.43 | 96.57 |
| | **NAA** | 91.75 | 96.39 | 95.88 | 95.36 | 95.36 | 93.81 | 94.33 |
| | **GN** | 95.32 | 94.15 | 90.06 | 92.40 | 94.15 | 95.91 | 98.83 |
| | **AM** | 99.49 | 98.98 | 100.00 | 100.00 | 99.49 | 99.49 | 100.00 |
| | **PCC** | 93.89 | 98.89 | 98.33 | 98.33 | 95.00 | 96.11 | 97.22 |
| **Pre.** | **ACA** | 81.13 | 69.01 | 79.03 | 79.03 | 80.36 | 84.91 | 89.29 |
| | **NAA** | 65.96 | 81.58 | 77.78 | 75.00 | 75.00 | 72.09 | 75.00 |
| | **GN** | 83.67 | 81.13 | 73.02 | 77.59 | 80.39 | 86.54 | 95.92 |
| | **AM** | 96.77 | 94.12 | 100.00 | 100.00 | 96.97 | 96.97 | 100.00 |
| | **PCC** | 77.55 | 93.94 | 92.11 | 93.02 | 83.02 | 85.42 | 89.36 |
| **F1** | **ACA** | 80.37 | 78.40 | 84.48 | 84.48 | 81.82 | 84.11 | 90.91 |
| | **NAA** | 75.61 | 84.93 | 78.87 | 76.06 | 76.06 | 79.49 | 83.54 |
| | **GN** | 76.64 | 77.48 | 76.03 | 77.59 | 75.23 | 81.82 | 87.85 |
| | **AM** | 93.75 | 95.52 | 95.24 | 95.24 | 96.97 | 96.97 | 100.00 |
| | **PCC** | 77.55 | 75.61 | 80.46 | 86.96 | 86.27 | 84.54 | 87.50 |

.

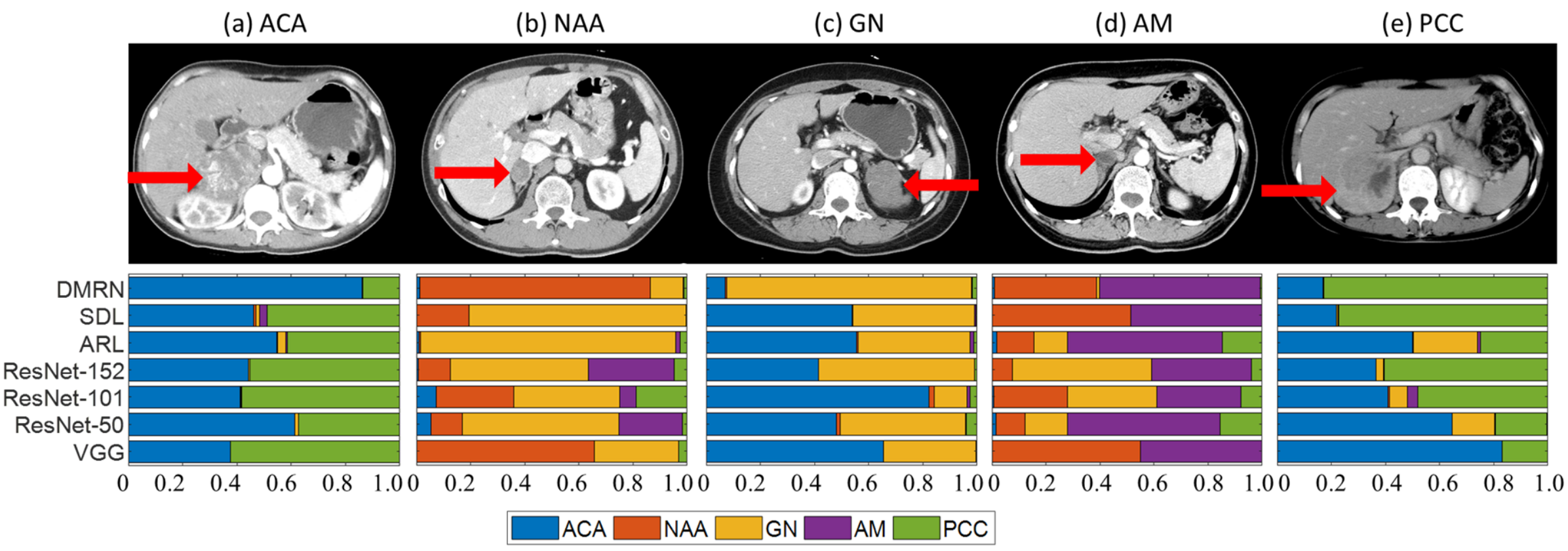

Figure 3. Classification results of five examples of adrenal masses with CT scans (top row) and results of different methods (columns). Red arrows indicate tumor locations, and color bars correspond to the probabilities derived from different methods for predicting individual sub-classes.

Table 4. Multi-class classification results on transaxial slices.

| | Acc. | Sen. | Spe. | Pre. | F1 | *p*-Value |
|---|---|---|---|---|---|---|
| **ResNet-50** | 77.64 | 72.14 | 93.39 | 76.26 | 73.58 | $p < 0.01$ |
| **ResNet-101** | 80.96 | 75.46 | 94.51 | 79.04 | 76.93 | $p < 0.01$ |
| **ResNet-152** | **81.85** | 75.75 | 94.79 | **79.29** | 77.33 | $p < 0.01$ |
| **ARL** | 79.48 | 77.11 | 94.21 | 76.76 | 76.57 | $p < 0.01$ |
| **SDL** | 81.11 | **79.28** | **94.80** | 76.72 | **77.52** | $p < 0.01$ |
| **DMRN** | **85.17** | **83.65** | **95.91** | **81.83** | **82.12** | - |

Table 5. Multi-class classification results on transaxial slices for individual sub-classes.

| | | ResNet-50 | ResNet-101 | ResNet-152 | ARL | SDL | DMRN (Ours) |
|---|---|---|---|---|---|---|---|
| **F1** | **ACA** | 84.26 | 86.30 | **88.02** | 86.83 | 86.72 | **91.01** |
| | **NAA** | 61.48 | **64.26** | 58.75 | 62.15 | 63.99 | **69.04** |
| | **GN** | 76.17 | **81.65** | 78.76 | 71.34 | 79.60 | **82.89** |
| | **AM** | 81.22 | 80.87 | 83.74 | **89.33** | 81.55 | **89.84** |
| | **PCC** | 64.75 | 71.58 | **77.41** | 73.23 | 75.74 | **77.83** |

### *3.2. Component Analysis*

We outline the results from individual stage of the DMRN in Table 6. Fig. 4 depicts the classification performance on individual classes. We note that when coupling random sampling (RS), multi-scale feature embedding (MS) with similarity feature learning (SFL), the feature representations have been greatly enhanced for differentiation, which resulted in higher classification accuracy.

Table 6. Classification accuracy with, and without, random sampling (RS), multi-scale feature embedding (MS) and similarity feature learning (SFL).

| | Acc. | Sen. | Spe. | Pre. | F1 |
|---|---|---|---|---|---|
| **DMRN (w.o. RS w.o. MS w.o. SFL)** | 79.04 | 78.93 | 94.67 | 79.24 | 79.00 |
| **DMRN (w.o. MS w.o. SFL)** | 81.22 | 80.48 | 95.24 | 81.00 | 80.51 |
| **DMRN (w.o. MS)** | 87.77 | 88.41 | 96.92 | 88.10 | 88.12 |
| **DMRN** | **89.52** | **90.73** | **97.34** | **89.91** | **89.96** |

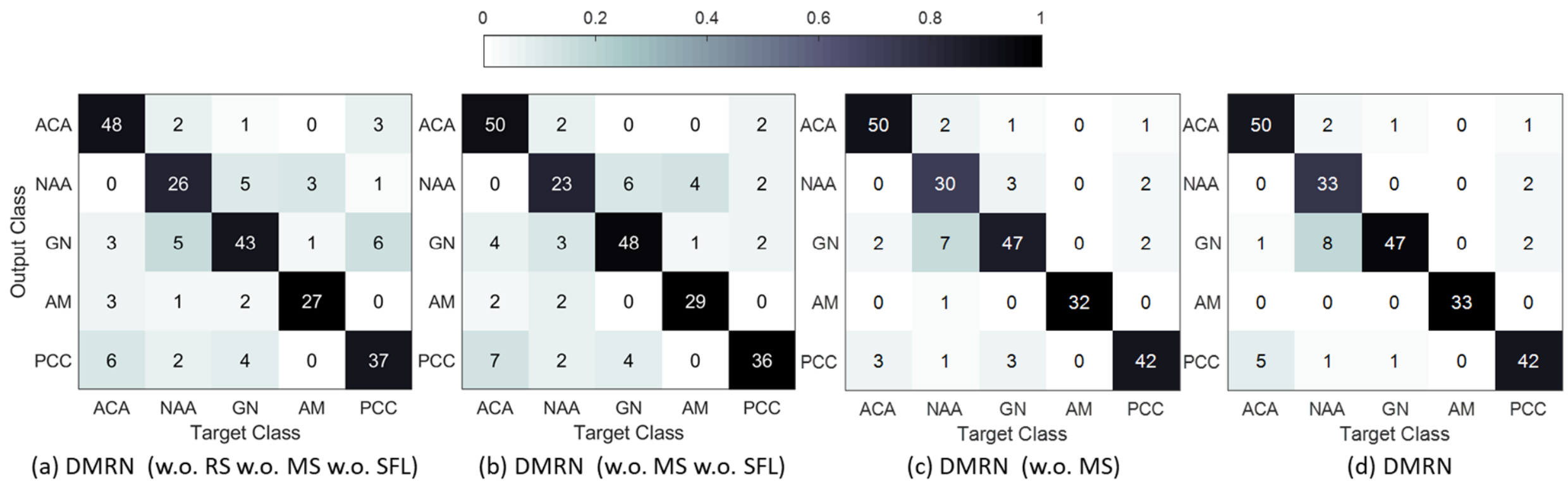

Figure 4. Confusion matrix of the classification results with, and without, random sampling (RS), multi-scale feature embedding (MS) and similarity feature learning (SFL).

### *3.3. Classification Results Analysis*

Fig. 5 is the feature visualization results derived using the t-SNE (t-distributed stochastic neighbor embedding) toolbox [41]. Image features were extracted from transaxial slices. t-SNE uses a nonlinear dimensionality reduction technique for visualizing high-dimensional image features in a low-dimensional space, and this can be used to indicate affinities (relationships) among different classes. Compared with ResNet-101, features derived from SDL and our DMRN presents a clear separation of different sub-classes in t-SNE visualization. ResNet-101 tended to overfit to the dominant classes (ACA and GN) and performed poorly in the rest of the classes. SDL had difficulties separating NAA, AM and GN. Our method, in contrast, retained a consistent better classification across different classes.

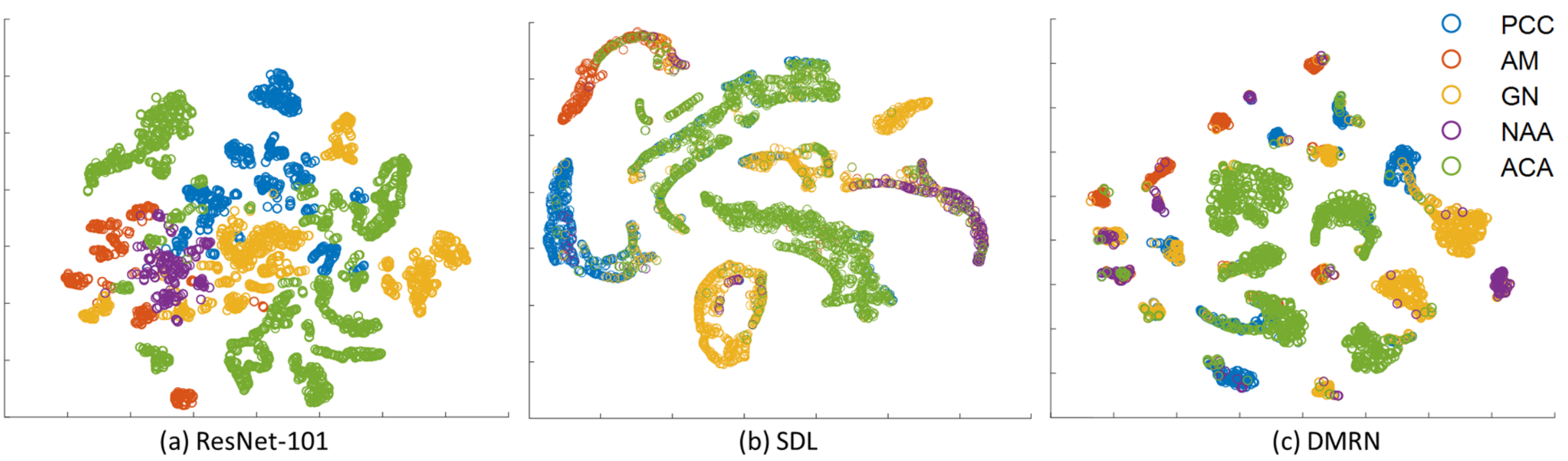

Figure 5. Feature embedding from (a) ResNet-101; (b) SDL; and (c) our DMRN methods, with t-SNE toolbox. Different color corresponds to different adrenal mass sub-classes. Features were extracted from transaxial slices.

### *3.4. Generalizability Analysis*

We conducted a generalizability analysis of our method on a second dataset. We used the medical Subfigure Classification dataset from the Image Conference and Labs of the Evaluation Forum (ImageCLEF) 2016 competition [42]. While a multitude of different types of images have been collected to assist in the development of more advanced CADs, the labelling of the collated image data remains problematic [43]. In cases where appropriate labels are absent, automatic identification of the imaging modality is an initial important step because semantics and content of an image can vary greatly depending on the modality. The ImageCLEF 2016 dataset is highly imbalanced across 30 different classes, as shown in Fig. 6. We used the standard pre-defined training set of 6,776 images and test set of 4,166 images. We used 90% (randomly selected) pre-defined training dataset as the training dataset and the remaining 10% as the validation dataset. The validation dataset was only used for tuning the parameters. The trained model was then evaluated on the pre-defined test dataset. We compared our method to the state-of-the-art methods reported for the dataset and this included: (1) ENS [44] – an ensemble of 15 different CNN models; (2) CDHVF [45] – a combined deep and handcrafted visual feature algorithm, where 3 CNN models together with speeded-up robust feature (SURF) and Local Binary Pattern (LBP) feature were used for classification; (3) GoogleNet+AlexNet [30] – an ensemble of GoogleNet with AlexNet; and (4) ResNet-152 [46].

Table 7 shows that our DMRN method had a better performance when compared to CDHVF, ResNet-152 and GoogleNet+AlexNet methods. The marginally degraded performance compared to the ENS method is likely due to ENS using additional training data (~2000 images). Our method had a competitive performance without using additional training data. We suggest that our method may further benefit through the addition of training data.

Table 8 shows the per class classification results of the proposed method. The results were measured with a F1 score and were compared with the GoogleNet+AlexNet method. It also shows the sum of the best results across different classes. Compared with GoogleNet+AlexNet method, our method has 21 classes achieved the best results, where GoogleNet+AlexNet only has 5 classes. This number indicates that our method achieved consistent better classification across different classes and was not overfit to the dominant classes.

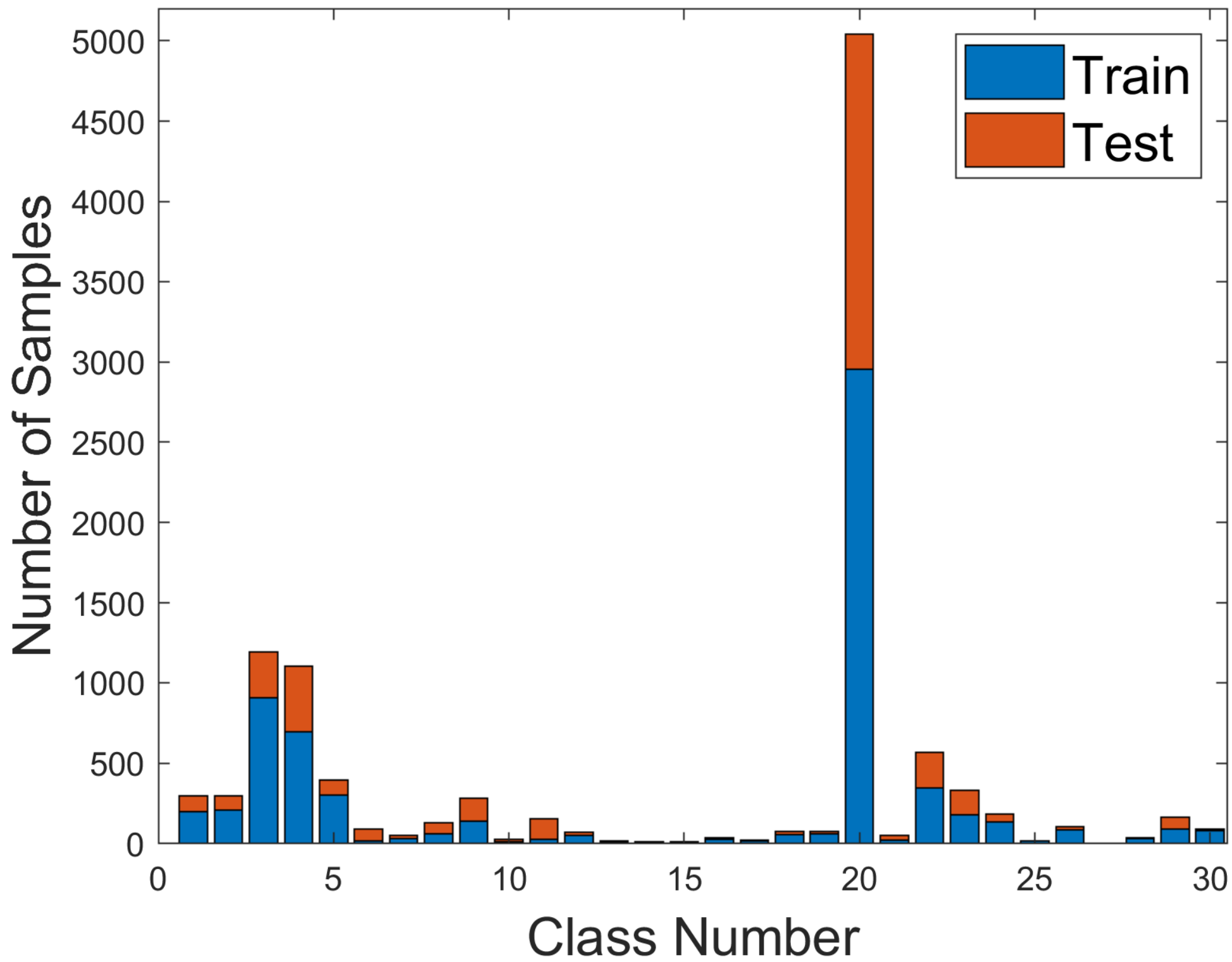

Figure 6. Distribution of the ImageCLEF dataset across different classes.

Table 7. Classification accuracy of our DMRN method with the existing methods on the ImageCLEF 2016 dataset. * indicates additional training data were used for model development.

| | **Acc.** |
|---|---|
| **GoogleNet+AlexNet** | 82.48 |
| **ResNet-152*** | 85.38 |
| **CDHVF** | 85.47 |
| **ENS*** | 87.37 |
| **DMRN** | 85.90 |

Table 8. Per class classification results measured by F1 score on the ImageCLEF 2016 dataset.

| Class # | Class | # Samples | | F1 | |
|---|---|---|---|---|---|
| | | Train | Test | GoogleNet+AlexNet | DMRN |
| 1 | 3D reconstructions (D3DR) | 201 | 94 | 71.07 | **79.79** |
| 2 | electron microscopy (DMEL) | 208 | 88 | 29.79 | **47.06** |
| 3 | fluorescence microscopy (DMFL) | 906 | 284 | 81.50 | **86.24** |
| 4 | light microscopy (DMLI) | 696 | 405 | 89.86 | **90.51** |
| 5 | transmission microscopy (DMTR) | 300 | 96 | 54.55 | **58.10** |
| 6 | angiography (DRAN) | 17 | 76 | 47.06 | **61.67** |
| 7 | combined modalities (DRCO) | 33 | 17 | 34.48 | **45.16** |
| 8 | computerised tomography (DRCT) | 61 | 71 | 82.99 | **84.35** |
| 9 | magnetic resonance (DRMR) | 139 | 144 | 82.39 | **84.14** |
| 10 | positron emission tomography (DRPE) | 14 | 15 | **23.53** | 22.22 |
| 11 | ultrasound (DRUS) | 26 | 129 | 70.00 | **88.41** |
| 12 | x-ray, 2D radiography (DRXR) | 51 | 18 | 44.00 | **48.28** |
| 13 | electrocardiography (DSEC) | 10 | 8 | 0 | 0 |
| 14 | electroencephalography (DSEE) | 8 | 3 | **100.0** | 0 |
| 15 | electromyography (DSEM) | 5 | 6 | 0 | 0 |
| 16 | dermatology, skin (DVDM) | 29 | 9 | 58.82 | **78.26** |
| 17 | endoscopy (DVEN) | 16 | 8 | **22.22** | 0 |
| 18 | other images (DVOR) | 55 | 21 | **56.52** | 53.33 |
| 19 | chemical structure (GCHE) | 61 | 14 | 92.86 | **96.30** |
| 20 | statistics, figures, graphs, charts (GFIG) | 2954 | 2085 | 93.73 | **95.53** |
| 21 | flowcharts (GFLO) | 20 | 31 | **26.32** | 12.12 |
| 22 | chromatography, gel (GGEL) | 344 | 224 | 84.94 | **86.64** |
| 23 | gene sequence (GGEN) | 179 | 150 | 33.67 | **46.03** |
| 24 | hand-drawn sketches (GHDR) | 136 | 49 | 30.77 | **40.45** |
| 25 | mathematics, formula (GMAT) | 15 | 3 | 0 | 0 |
| 26 | non-clinical photos (GNCP) | 88 | 20 | 35.90 | **51.16** |
| 27 | program listing (GPLI) | 1 | 2 | 0 | 0 |
| 28 | screenshots (GSCR) | 33 | 6 | 25.00 | **28.57** |
| 29 | system overviews (GSYS) | 91 | 75 | 11.11 | **32.47** |
| 30 | tables and forms (GTAB) | 79 | 13 | 48.00 | **69.23** |
| **Best Results Count** | | | | #5 | **#21** |

## 4. DISCUSSION

Our main findings are that: (1) our DMRN had the highest accuracy in the classification of adrenal masses; (2) the features derived from our method were tolerant to large intra-class variations and could separate the subtle differences between sub-classes; and (3) the paired inputs ensured that all sub-classes had equal contributions to the trained model and the risk of overfitting to the dominant classes was minimized.

Our DMRN outperformed the other methods that were evaluated ($p < 0.05$). The SDL was the second-best performed; it also used multiple CNNs to learn competitively as did our approach. The reliance on using cross-entropy loss meant that all the training samples had the same impact on the feature learning outcomes. We also modeled the similarities and dissimilarities based on the data correlation (feature distance) among paired training samples. Hence, our DMRN adapted to the easy training samples, and then gradually adapted to the more difficult samples. Whilst the SDL used a single level feature learning process; we employed multi-scale feature embedding so that our feature descriptors were trained with multi-level supervision. As a result, our feature descriptors were more descriptive at both low-level appearance information and high-level semantic information. In the challenging cases, the SDL-derived features had difficulty in identifying sub-classes.

We showed that different adrenal masses could be separated based on the features extracted via the DMRN. In contrast, features derived from the baseline ResNet-101 did not provide a clear separation. Features derived from SDL had difficulty separating the challenging sub-classes. We attribute the better performance of our method to the similarity feature learning that used similarity loss to group neighbors from the same sub-classes while retaining a large margin for samples from different sub-classes.

Our training dataset was also imbalanced and the number of transaxial image slices for adrenocortical carcinomas (ACAs) was seven times greater than non-functional adrenal adenomas (NAAs) and four times greater than adrenal myelolipomas (AMs). ResNet-101 tended to overfit to the dominant classes and had problems classifying the other subtypes especially NAA and AMs. In contrast, DMRN used paired data for training so data from different classes were sampled equally and so minimized the impact of imbalanced training data. The differences between HC and VGG shows the benefit of using CNNs for feature extraction. The deep learning hierarchical structure allowed the derivation of deep semantic representations and resulted in higher classification accuracy. The improvement of the VGG method over 3D-CNN is expected. For a CNN network with the same number of convolutional layers, 3D-CNN requires more GPU memory compared to a 2D- based CNN. The poor performance of the 3D-CNN was likely due to the limited size of the GPU memory which restricted the number of convolutional layers to be inserted into the 3D networks. Consequently, the learned features from the 3D-CNN will be sub-optimal for differentiating the different adrenal masses.

The improvement of ResNet when compared to the VGG, we suggest, is likely to be due to using residual blocks that allowed an increase in the overall depth of the network. When compared with ResNet based methods such as ResNet-50 and ResNet-101, ARL showed a marginal 1% improvement in accuracy. ARL uses an attention module to focus on subtle inter-class features which minimizes the problem of large inter-class similarities. However, without modelling intra-class variations, it will be problematic for an attention module to group adrenal masses with large intra-class variations into the same sub-classes.

### *4.1. Limitations and Future Work*

With the express purpose of illustrating the performance of multi-scale resemblance network, we have not optimized the architecture of the encoder, but rather chose to use the well-established ResNet encoder models. More recent encoders such as models based on transformer networks may provide better performance, albeit with a much greater computational expense and GPU memory consumption. The extension of the proposed method to be in a transformer network is a possible option for further investigation.

In our experiments, we focused on classifying the identified adrenal lesions into different sub-classes. Identification of such lesions require a manual input as the pre-processing step. To make the process more streamlined and applicable to clinical workflow, an initial automatic lesion detection may be applied to detect the bounding boxes that contain lesions.

Based on the superior generalizability of our method, we will investigate the adaptations of our method to support other medical image datasets which have similar large intra-class variations, large inter-class similarities and imbalanced training data problem such as in skin lesion classification in dermoscopy images. We also plan on exploring the adaptation of our method to different adrenal lesion applications, including automatic adrenal lesion monitoring over multiple image time-points.

## 5. Conclusions

We proposed a CAD approach to identify different adrenal masses and we showed that our deep multi-scale resemblance network had better accuracy ($p < 0.05$) when compared to state-of-the-art methods. We suggest that our method could be used for the diagnostic separation of different adrenal masses.

## Acknowledgement

This work was supported in part by Australian Research Council (ARC) grants (DP200103748 and IC170100022).